%% file: main.tex
\documentclass[10pt,twocolumn,letterpaper]{article}

\usepackage{iccv}              


\usepackage[accsupp]{axessibility} 
\definecolor{iccvblue}{rgb}{0.21,0.49,0.74}
\usepackage[pagebackref,breaklinks,colorlinks,allcolors=iccvblue]{hyperref}
\usepackage{times}
\usepackage{epsfig}
\usepackage{graphicx}
\usepackage{amsmath}
\usepackage{amssymb}
\usepackage{caption}
\usepackage{booktabs}
\usepackage{multirow}
\usepackage{colortbl}
\usepackage{soul}
\usepackage{tablefootnote}
\usepackage{xcolor}
\usepackage{bbding} 

\def\paperID{361} 
\def\confName{ICCV}
\def\confYear{2025}

\title{IRGPT: Understanding Real-world Infrared Image with Bi-cross-modal Curriculum on Large-scale Benchmark}

\author{
\begin{tabular}{ccc}
Zhe Cao & Jin Zhang & Ruiheng Zhang\thanks{Corresponding author} \\
Beijing Institute of Technology & Beijing Institute of Technology & Beijing Institute of Technology \\
Beijing, China & Beijing, China & Beijing, China \\
{\tt\small zhe.cao@bit.edu.cn} & {\tt\small jinzhang@bit.edu.cn} & {\tt\small ruiheng.zhang@bit.edu.cn} \\
\end{tabular}
}

\begin{document}
\maketitle

\renewcommand\twocolumn[1][]{#1}%
\maketitle


\input{sec/0_abstract}    
\input{sec/1_intro}

{
    \small
    \bibliographystyle{ieeenat_fullname}
    \bibliography{main}
}

\end{document}

%% file: sec/0_abstract.tex
\begin{abstract}
    Real-world infrared imagery presents unique challenges for vision-language models due to the scarcity of aligned text data and domain-specific characteristics. 
    Although existing methods have advanced the field, their reliance on synthetic infrared images generated through style transfer from visible images, which limits their ability to capture the unique characteristics of the infrared modality.
    To address this, we propose IRGPT, the first multi-modal large language model for real-world infrared images, built upon a large-scale InfraRed-Text Dataset (IR-TD) comprising over 260K authentic image-text pairs. 
    The proposed IR-TD dataset contains real infrared images paired with meticulously handcrafted texts, where the initial drafts originated from two complementary processes: (1) LLM-generated descriptions of visible images, and (2) rule-based descriptions of annotations. 
    Furthermore, we introduce a bi-cross-modal curriculum transfer learning strategy that systematically transfers knowledge from visible to infrared domains by considering the difficulty scores of both infrared-visible and infrared-text. Evaluated on a benchmark of 9 tasks (e.g., recognition, grounding), IRGPT achieves state-of-the-art performance even compared with larger-scale models.
\end{abstract}

%% file: sec/1_intro.tex
\section{Introduction}

\begin{figure}
    \centering
    \includegraphics[width=0.86\linewidth]{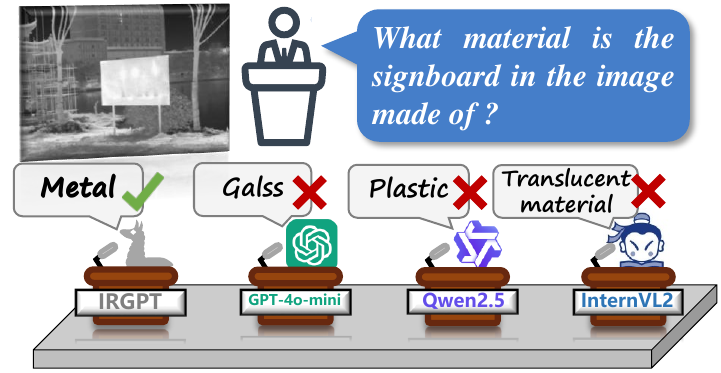}
    \caption{Exist MLLMs exhibit hallucinations when interpreting real-world infrared images, while IRGPT effectively suppresses this phenomenon.}
    \label{Figure0}
\end{figure}

With the advancement of large model technologies, Vision-Language Models (VLMs) designed for visible images have achieved significant breakthroughs \cite{zhang2024vision, li2024seed}. This success can be attributed to two main factors: the availability of large-scale aligned image-text datasets \cite{schuhmann2022laion, ikezogwo2023quilt} and the powerful visual perception capabilities provided by pre-trained vision encoders \cite{dosovitskiy2020image, kirillov2023segment} for Large Language Models (LLMs). VLMs such as LLaVA \cite{LLaVA} and Qwen \cite{yang2024qwen2} 
have demonstrated exceptional vision capabilities, enabling them to deeply understand real-world natural images through the medium of corresponding text. However, infrared images, a distinctive and widely utilized image type, have been overlooked by multi-modal large language models (MLLMs), leading to frequent hallucinations when these models process infrared imagery. As illustrated in \cref{Figure0}, the salient metallic signboard in the image is erroneously identified as a non-metallic material object.

\begin{figure*}
    \centering
    \includegraphics[width=0.85\linewidth]{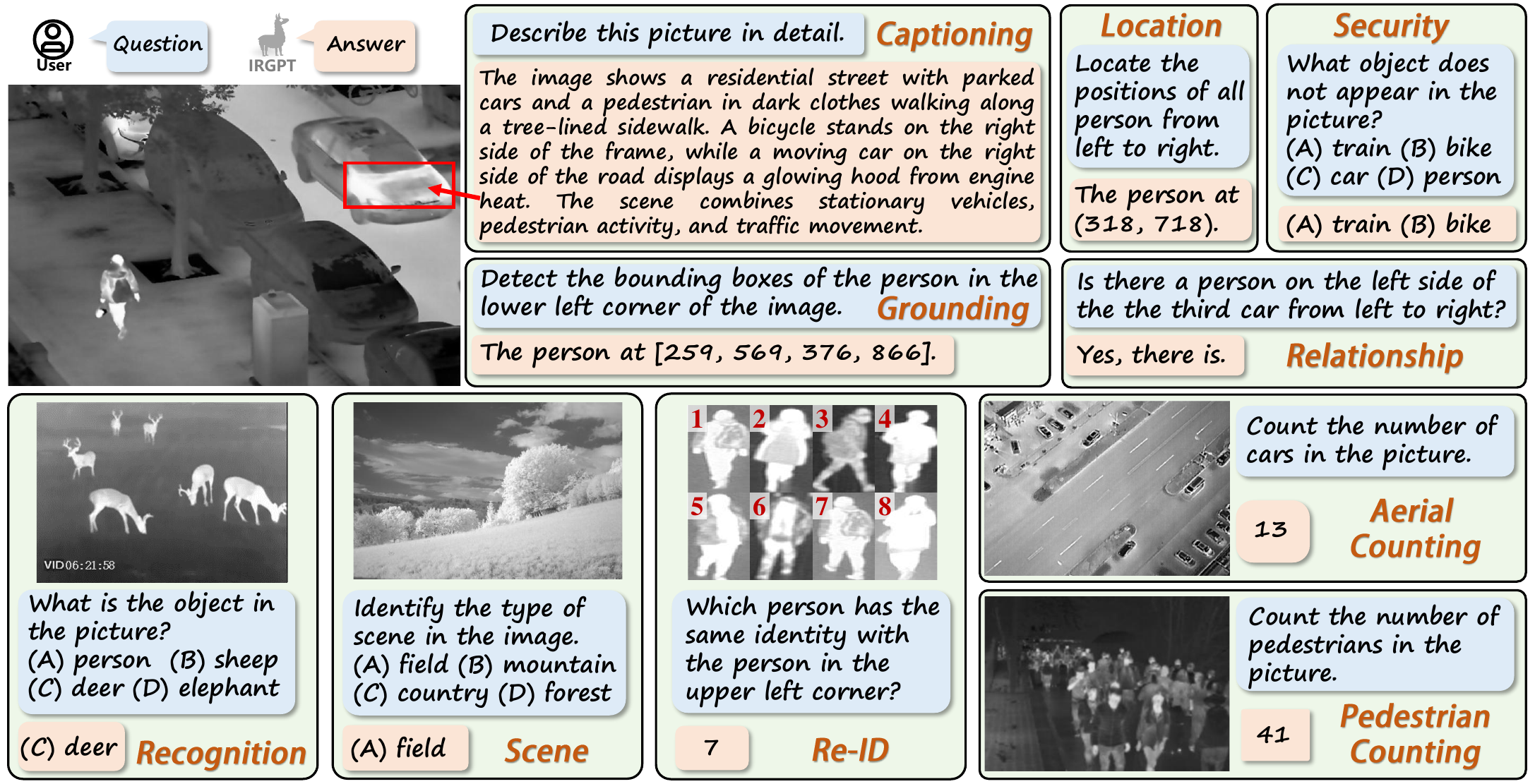}
    \caption{IRGPT effectively handles complex infrared image-text tasks (e.g., grounding, recognition), demonstrating successful knowledge transfer from visible to infrared domains and a deep semantic understanding of infrared data. We further showcase IRGPT's captioning capability in understanding that moving vehicles exhibit elevated engine temperatures.} 
    \label{fig1}
\end{figure*}

In the field of infrared images, the scarcity of large-scale image-text datasets poses challenges to the development of infrared MLLMs. Prior studies \cite{jiang2024infrared, girdhar2023imagebind} have explored methods to obtain infrared-text paired data. However, these methods using visible images to synthesize infrared counterparts create a modality gap, as the synthesised data quality directly depends on generative models. Moreover, such data strictly adheres to the training distribution of the generative model, neglecting the authenticity of infrared images.

To establish a realistic and effective infrared-text dataset (IR-TD) for advancing infrared MLLMs, we also leverage visible images as a medium to employ a new-path approach to obtain infrared-text pairs. On the one hand, we manually choose visible-infrared image pairs and utilize LLMs to generate draft descriptions for visible images, which are then carefully adapted to annotate corresponding infrared images and create Q\&A pairs. On the other hand, we develop rule-based methods for caption generation and Q\&A pair construction, specifically designed by various annotations. The resulting IR-TD, containing 260K+ high-quality infrared-text pairs, adheres to two fundamental principles: 1) authenticity of infrared imagery, and 2) semantic alignment between modalities. 

While addressing the data scarcity issue, IR-TD introduces new challenges stemming from inherent characteristics of real-world infrared images, including sparse semantic information, imaging disparities across different spectral bands, and thermal information's dominance in target saliency. 
These factors render it a significant challenge for an infrared-specific MLLM developed from scratch to rival those MLLMs extensively trained on visible images.


To address this challenge, we introduce a bi-cross-modal curriculum that builds upon the robust comprehension capabilities of existing MLLMs. By leveraging the similarity between infrared and visible images, as well as the semantic alignment between infrared images and text, our approach transfers MLLM understanding from visible to real-world infrared image through incremental pre-training from simple to complex infrared data. With this approach, we propose \textbf{IRGPT}, the first infrared-specific MLLM grounded in real-world infrared image.
As illustrated in \cref{fig1}, the developed model showcases strong cross-modal understanding capabilities, particularly in infrared question-answering tasks. To validate IRGPT's performance on common infrared tasks, we construct a benchmark with 9 different infrared Q\&A tasks containing 37K+ test samples on the IR-TD dataset. Through evaluations across various tasks, our model achieves the state-of-the-art performance, demonstrating significant potential for practical applications. The main contributions of this paper are as follows:

\begin{itemize}
    \item We introduce a large-scale open-source infrared-text dataset, comprising pre-trained subset, fine-tuning instruction subset, and a multi-task benchmark for infrared-text data, which includes 9 types of tasks such as recognition and localization. The total number of samples exceeds 260K.

    \item To the best of our knowledge, we propose the first vision-language model based on real-world infrared images named IRGPT, for which both the incrementally pre-trained and fine-tuned infrared images are collected from the real world rather than being synthetically generated.

    \item We also propose a bi-cross-modal curriculum transfer learning strategy that balances the transferability between infrared and visible images and the semantic consistency between infrared images and text.
    
    \item In the 9 benchmark tasks, our IRGPT achieves state-of-the-art (SOTA) results. Under zero-shot settings, IRGPT demonstrates a 76.35 improvement in positive sum and a 31.56 reduction in negative sum compared to baseline.

\end{itemize}

\section{Related Works}
\textbf{Multi-modal Large Language Model.}
In the real world, information exists in diverse forms, making it impossible for a simple LLM to perceive and understand the world fully. Significant efforts have been made to explore how LLMs can adapt to multi-modal data \cite{zhu2024minigpt, laurenccon2025matters}, with notable advancements such as the LLaVA framework \cite{LLaVA}, which has enabled LLMs to ``see" and interpret visual information. From the perspective of current advancements in state-of-the-art MLLMs \cite{li2023blip,alayrac2022flamingo,chen2024internvl,wang2024qwen2,wu2024deepseek}, supporting visual modality data has become an inevitable trend. However, most of these models overlook the infrared images with unique characteristics \cite{caffagni2024r}. Treating infrared images as equivalent to visible images can not avoid leading to severe hallucinations, posing significant challenges for downstream applications and deep understanding of the real scenario.

\textbf{Infrared Image in Large Models.}
With the development of multi-modal foundation models, infrared image processing has seen two primary research directions. (a) Unimodal approaches focus on intrinsic infrared characteristics: InfMAE \cite{liu2025infmae} employs masked image modeling for foundational model training, while IRSAM \cite{zhang2025irsam} adapts SAM \cite{kirillov2023segment} for small target detection through edge enhancement. (b) Multi-modal methods leverage cross-modal synergies: TC-MoA \cite{zhu2024task} uses MoE frameworks for modality fusion, and works like LanguageBind \cite{zhulanguagebind}, PandaGPT \cite{su2023pandagpt}, and ImageBind \cite{girdhar2023imagebind} connect thermal infrared with text/visible modalities.
The most relevant work is Infrared-LLaVA \cite{jiang2024infrared}, which trains an infrared-text large model using the data generated by sRGB-TIR \cite{sRGB-TIR}. However, its reliance on video-generated images, which lack authenticity, negatively introduces modality-related hallucinations \cite{liu2024survey}.

\textbf{Multi-modal Curriculum Learning.}
Curriculum learning (CL), first proposed in \cite{Bengio2009Curriculum}, mimics human developmental learning by progressively training models from simple to complex samples through data sequencing strategies \cite{soviany2022curriculum,wang2021survey}. This approach has proven effective for enhancing both training efficiency and model generalization in conventional deep learning \cite{wang2023efficienttrain, li2022stability, karim2023c}.
Our analysis highlights two critical CL directions requiring deeper exploration: 1) its application in large-scale models, and 2) curriculum design for multi-modal learning \cite{xu-etal-2020-curriculum,liu2024let, xi2024training, wang2020learning}. Recent work shows initial progress in addressing multi-modal training cases \cite{NEURIPS2023_a1e6783e, zhou2023intra}, while existing methods still lack explorations for training multi-modal large-scale models effectively.


\begin{figure}
    \centering
    \includegraphics[width=0.96\linewidth]{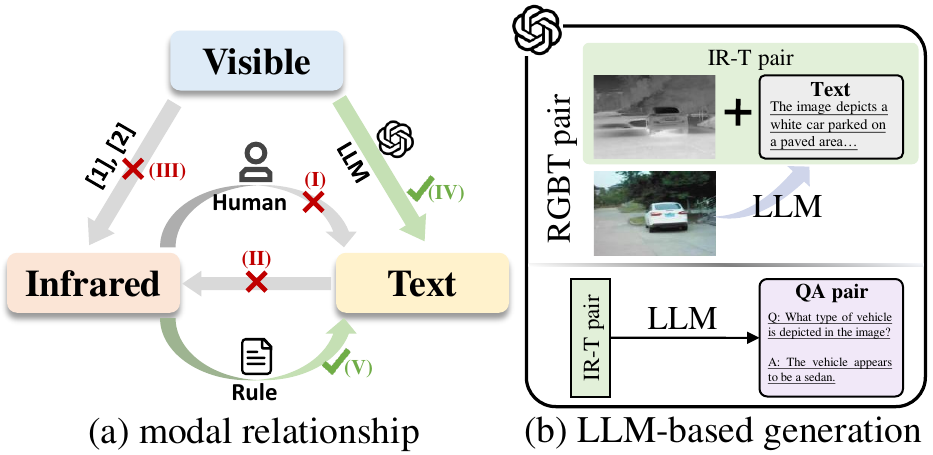}
    \caption{The translation of tri-modal data. (a) Tri-modal data transformation relationships and potential generative paths. (b) The LLM-based infrared-text pair generation method in IR-TD.}
    \label{fig:Generation}
\end{figure}
\section{IR-TD: Infrared-Text Dataset}

\subsection{Tri-modal Data Translation}

In the field of infrared imaging, large-scale text annotation shows a distinct scarcity \cite{liu2020lsotb, liu2025infmae}. On one hand, textual descriptions for infrared images cannot be effectively obtained through web scraping like the visible image. On the other hand, the difficulty of describing infrared images seriously affects the quality of annotation. As a result, conventional methods for image-text pair collection cannot be directly applied to the infrared domain. With recent advancements in generative models, some works (e.g., IID \cite{jiang2024infrared}) have attempted to synthesize infrared-text pairs through style transfer approaches. Such methods adhere to fixed generative distributions, which severely degrade the quality of data reliant on synthetic models. To explore pathways for generating infrared-text pairs, we analyze the feasibility of transformations between infrared images, visible images, and textual descriptions, as illustrated in \cref{fig:Generation}(a).

\textbf{(I) Manual Annotation}: This approach faces significant challenges, particularly the difficulty in achieving diversity, making it impractical for large-scale data pair collection.

\textbf{(II) Text-to-Image Generation}: Most models do not support the generation of infrared data.

\textbf{(III) Synthetic Image Generation}: Converting visible images into infrared ones fails to preserve the authenticity.

\textbf{(IV) Generation with Aligned RGBT Pairs}: Leveraging LLMs to describe visible images that correspond to infrared images and using these descriptions as proxies for infrared image annotations.

\textbf{(V) Rule-Based Generation}: According to the labels, it is possible to generate large-scale image-text pairs, but the drawback is the lack of diversity.

Our proposed solution integrates approaches (IV) and (V) to balance quality and diversity. Specifically, for infrared-visible paired images, we utilize LLMs to generate detailed descriptions of visible images, which are then adopted as textual annotations for the corresponding infrared images. For special scenarios where visible image is insufficient (e.g., nighttime conditions), we employ Retinexformer \cite{cai2023retinexformer} to improve observability, which can be found in Appendix B. In other cases where visible images present challenges, such as small targets or camouflaged objects, we employ a rule-based approach to generate image-text pairs with available annotations, ensuring that the constructed IR-TD has correct and clear semantics.
To effectively train infrared MLLMs, we partition IR-TD into three distinct segments: the pre-training subset for incremental pre-training, the instruction subset for fine-tuning, and the benchmark for validation. \cref{tab:dataset} provides some fundamental characteristics of our proposed IR-TD, in comparison with the closed-source dataset IID in \cite{jiang2024infrared}.

\subsection{Pre-training Subset Generation}\label{pretrain set}
We aggregated 63 publicly available datasets (details in Appendix H) to collect raw RGBT pairs. To address field-of-view discrepancies caused by inconsistent camera parameters (e.g., datasets like \cite{UVT2000, UVT20K}), we aligned semantic content by cropping images based on object locations derived from labels, with examples provided in Appendix J. This ensured textual consistency between infrared-visible pairs. We also rigorously resampled datasets to mitigate redundancy from video-extracted infrared frames (e.g., retaining only 1$\%$ of VTUAV's 1.7M images). The final curated dataset contains 84,284 RGBT pairs. As shown in \cref{fig:Generation}(b), visible images were processed by LLMs to generate descriptive texts, while rule-based methods created 106k infrared-text pairs using annotations. 

\begin{table}[t]
  \centering
  \resizebox{0.9\linewidth}{!}{
    \begin{tabular}{l|cc}
    \toprule
         \textbf{Dataset} & \textbf{IID\cite{jiang2024infrared}} & \textbf{IR-TD\tablefootnote{\href{https://github.com/WheatCao/ICCV2025-IRGPT}{The dataset is available here.}}} \\
    \midrule
    \textbf{Pre-training image samples} & 118k  & 190k \\
    \textbf{Instruction samples} & 12k   & 33k \\
    \textbf{Benchmark samples} & 22k   & 37k \\
    \textbf{Original datasets} & 7     & 63 \\
    \textbf{Benchmark tasks} & 6     & 9 \\
    \midrule
    \textbf{Open-source} & \XSolidBrush  &  \Checkmark  \\
    \textbf{Image authenticity} & \XSolidBrush  & \Checkmark \\
    \textbf{Data Redundancy} &  \Checkmark  & \XSolidBrush \\
    \bottomrule
    \end{tabular}}
  \caption{The fundamental characteristics of the IR-TD, along with a comparative analysis against the closed-source IID in \cite{jiang2024infrared}.}
  \label{tab:dataset}%
\end{table}

\subsection{Instruction Subset Generation}
The fine-tuning instruction subset, similar to the pre-training subset, is bifurcated into LLM-based and rule-based components. The LLM-based portion employs the same prompts as LLaVA \cite{LLaVA} and generates Q$\&$A pairs based on the descriptive image-text pairs obtained in \cref{pretrain set}. This approach facilitates the effective transference of the LLM's question-answering capabilities from visible images to infrared images. Conversely, the rule-based segment is predominantly task-oriented, crafting Q$\&$A pairs that align with the task's framework to bolster the LLM's relevance in addressing task-specific queries when confronted with infrared images.

\subsection{Benchmark Defination} \label{Benchmark}
The benchmark primarily encompasses 9 cross-modal tasks, designed to comprehensively evaluate the various capabilities of the model. Details on the establishment of the benchmark can be found in the Appendix A. Specifically, we provide the following definitions:
\begin{itemize}

\item \textbf{Recognition}: Construct four options, one of which represents the target category contained in the image, and require the model to identify the correct target option.

\item \textbf{Grounding}: Provide a textual description specifying a target in the image and let model return its bounding box.

\item \textbf{Location}: Use text to specify a target category, and require the model to provide the coordinate locations of all targets of that category in the image.

\item \textbf{Relationship}: Determine whether the spatial relationship described in the text is correct.

\item \textbf{Re-ID}: Construct an image using eight samples arranged in a 2x4 grid, numbered 1-8. Only one sample matches the identity of the top-left sample, testing the model's re-identification capability.

\item \textbf{Security}: Construct four options, and require the model to identify all targets not present in the image.

\item \textbf{Aerial Counting}: Vehicle counting via UAV-captured overhead imagery in roadway and parking lot scenarios.

\item \textbf{Pedestrian Counting}: The image contains a large, crowded group of pedestrians, and the task requires estimating the number of pedestrians.

\item \textbf{Scene}: Construct four options, one of which correctly describes the scene in the image, and require the model to select the correct option.
\end{itemize}

\begin{figure}
    \centering
    \includegraphics[width=0.99\linewidth]{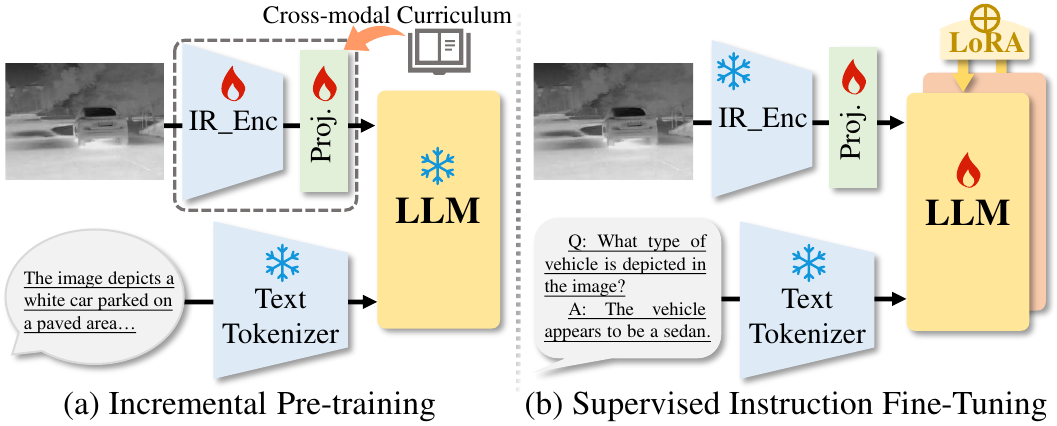}
    \caption{Training strategy of IRGPT. (a) Incremental pre-training stage trains the vision encoder and projector layer via curriculum learning for image-text alignment. (b) Supervised instruction fine-tuning adapts LLM with LoRA and also trains the projector.}
    \label{fig:train}
\end{figure}

\section{IRGPT: Infrared Vision-Language Model}\label{Curriculum Schedule}

\begin{figure*}
    \centering
    \includegraphics[width=0.85\linewidth]{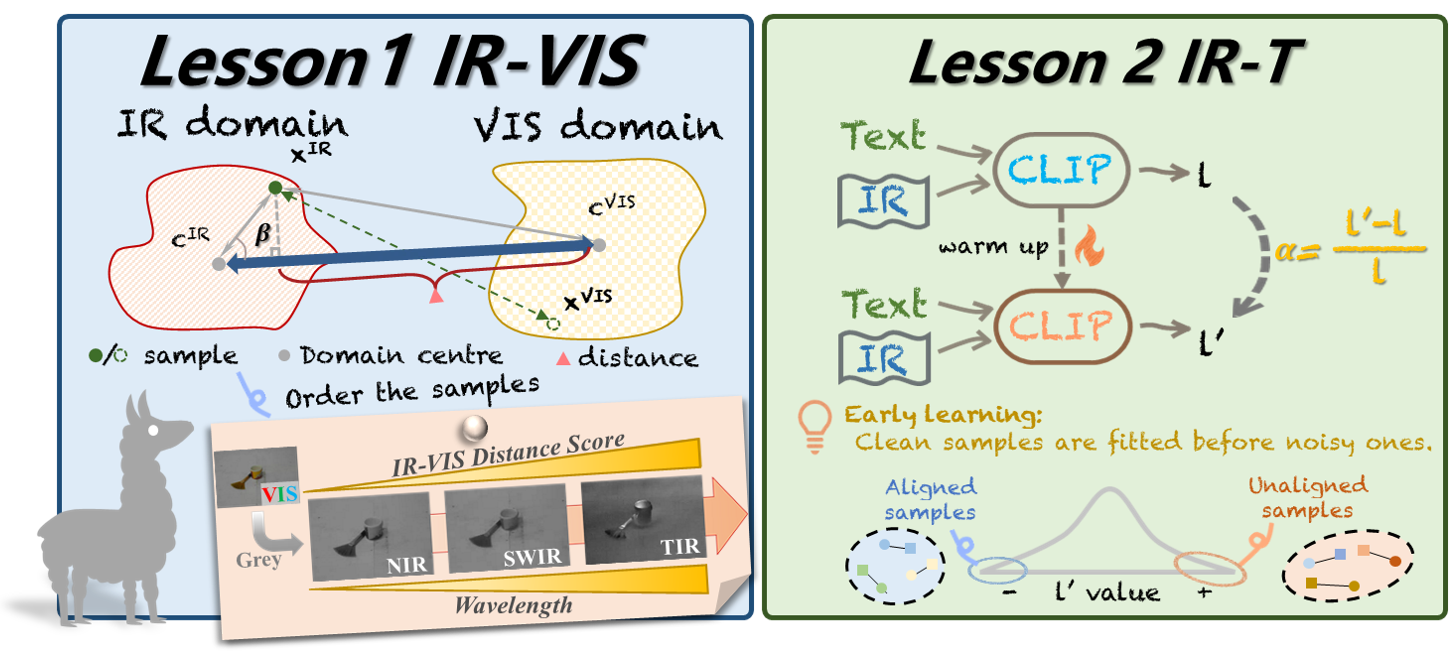}
    \caption{Bi-cross-modal curriculum transfer strategy. \textbf{Lesson 1} focuses on quantifying the domain gap between infrared and visible images through geometric metrics, constructing a domain projection distance as a sample transfer difficulty score. \textbf{Lesson 2} emphasizes semantic alignment between infrared images and text pairs, utilizing a pre-warmed CLIP to compute loss, which serves as a sample alignment difficulty score. Notably, the loss variation rate is treated as an auxiliary weight in pre-train. NIR: Near infrared, SWIR: Short wavelength infrared, TIR: Thermal infrared.}
    \label{fig:curriculum}
\end{figure*}

\subsection{Model Architecture and Training Strategy}
 
Even with the foundation of IR-TD, it is challenging to train a large-scale model dedicated to infrared images from scratch. This difficulty arises primarily due to two factors: absence of a vision-language pre-trained infrared encoder and the semantic sparsity inherent in infrared imagery (e.g. low contrast leading to less informative images). Consequently, our IRGPT will leverage a pre-trained MLLM as a foundation for transfer learning, enhancing its image comprehension and textual expression capability.
Our model consists primarily of a vision encoder, a base LLM, a text tokenizer, and a trainable vision projector. As illustrated in \cref{fig:train}, our training strategy is divided into incremental pre-training and supervised instruction fine-tuning.

\textbf{Incremental Pre-training}: This phase focuses solely on training the vision encoder and its corresponding projection layer. The objective is to adapt the vision encoder to effectively process infrared image inputs while ensuring alignment with textual data.

\textbf{Supervised Instruction Fine-tuning}: During this phase, both the visual projector and the LLM undergo fine-tuning. Notably, the LLM is fine-tuned with the Low-Rank Adaptation strategy to prevent the loss of expressive power.

The incremental pre-training process faces challenges due to the inherent complexity of training data, particularly the varying learning difficulties of infrared-text pairs, which significantly constrain model quality. To address this, we propose Bi-cross-modal Curriculum Transfer Learning, a novel framework that facilitates progressive knowledge transfer from visible to infrared image understanding.

\subsection{Bi-cross-modal Curriculum Transfer Learning}

\subsubsection{Lesson 1: IR-VIS}
The modality gap between infrared and visible images represents a critical component of sample difficulty. From an intuitive perspective, we propose to rank infrared images based on geometric metrics since images more similar to visible counterparts would be easier to learn for the vision encoder already adapted to visible images \cite{NEURIPS2020_0607f4c7,9134370Transfer}. Notably, our observations reveal that infrared images become progressively dissimilar from visible images with increasing detection wavelengths. Near-infrared images resemble grayscale versions of visible images, while thermal infrared images exhibit fundamentally different characteristics compared to visible images.

Our methodology employs a feature extractor in \cite{liu2025infmae} re-trained with masking strategies on both infrared and visible grayscale images to extract discriminative features. To achieve sample-level distance scoring, individual distance computation is required. However, the absence of paired visible counterparts for many infrared images necessitates an innovative solution. As shown in \cref{fig:curriculum}, we therefore propose calculating cross-modal distances through intra-domain distance projection, where the distance from each sample to the opposite modality is estimated by projecting intra-domain distances onto the inter-domain direction.

We first compute the domain discrepancy between infrared and visible domains using the widely adopted Maximum Mean Discrepancy (MMD) \cite{li2018domain, 9478936, ge2023unsupervised, du2021cross}:

\begin{equation}
\small    \mathrm{MMD}^2=\left\|\frac{1}{n} \sum_{i=1}^n \phi\left(x_i^{i r}\right)-\frac{1}{m} \sum_{j=1}^m \phi\left(x_j^{v i s}\right)\right\|_{\mathcal{H}}^2,
\end{equation}
where $\|\cdot\|_{\mathcal{H}}^2$ represents the squared norm in the Reproducing Kernel Hilbert Space (RKHS). $\phi: \mathcal{X} \to \mathcal{H}$ denotes the feature map induced by Gaussian kernel. $x_i^{ir} \in \mathcal{X}^{ir}$ and $x_j^{vis} \in \mathcal{X}^{vis}$ denote the $i$-th infrared domain sample and $j$-th visible domain sample, respectively.

Subsequently, we calculate domain center vectors:
\begin{equation}
\small c^{i r}=\frac{1}{n} \sum_{i=1}^n \phi\left(x_i^{i r}\right), \quad c^{v i s}=\frac{1}{m} \sum_{j=1}^m \phi\left(x_j^{v i s}\right).
\end{equation}

Finally, the sample distance score is obtained by projecting the infrared sample's distance vector to the infrared domain center onto the inter-domain direction and adding the domain discrepancy:

\begin{equation}
\small d_i=\left(\phi\left(x_i^{i r}\right)-c^{i r}\right) \cdot \frac{c^{v i s}-c^{i r}}{\left\|c^{v i s}-c^{i r}\right\|}+\mathrm{MMD}.
\end{equation}

This approach enables effective quantification of cross-modal sample difficulty while addressing the challenge of incomplete visible-infrared correspondence. It is worth noting that the distances obtained by this method are not precise. However, since our goal is to acquire a difficulty ranking of the samples, this approximation is acceptable.

\subsubsection{Lesson 2: IR-T}
Another critical factor constraining the visual comprehension capabilities of MLLMs lies in the alignment quality between visual and textual modalities. Empirical observations \cite{radford2021learning,conneau2019cross} suggest that semantically well-aligned samples demonstrate superior learnability, whereas premature exposure to loosely-aligned samples during training may induce oscillatory optimization trajectories that impede effective model convergence. Consequently, the degree of cross-modal alignment serves as an essential indicator for assessing sample complexity. Prior research \cite{huang2021learning,yang2023bicro} has investigated sample-level assessment of multi-modal alignment relationships, predominantly grounded in early learning principles. This theoretical framework posits that deep models inherently prefer assimilating clean samples over noisy counterparts during initial training phases.



Inspired by these insights, we implement a pre-warmed CLIP \cite{radford2021learning} to establish a discriminative criterion for infrared-text alignment through its loss values. However, the observed loss distribution displays pronounced Gaussian characteristics (as shown in \cref{fig:distribution}) with significant concentration around the mean value. This phenomenon obscures the identification of two crucial sample groups. The first group includes hard samples that require intensive optimization efforts, and the second consists of misaligned samples that warrant weight attenuation to prevent erroneous gradient updates. These issues stem from the static loss measurement from the pre-warmed model, which merely reflects instantaneous model performance and fails to capture dynamic learning behaviors.

To address the limitations posed by static loss, we propose a novel dynamic metric called loss variation rate. This metric is designed to dynamically assess model learning progress and optimize training efficiency by calculating the rate of change in loss as follows:


\begin{equation}\label{loss variation}
{\small \alpha = \frac{l'-l}{l},}
\end{equation}
where $l$ and $l'$ denote pre-warm-up and post-warm-up loss values respectively. This adaptive ratio enables dual functionality: 1) suppressing the influence of misaligned samples through weight modulation, while 2) preserving optimization focus on valuable hard samples. 

 \begin{table*}[ht]
  \centering
  \resizebox{0.9\linewidth}{!}{
    \begin{tabular}{l|cccccc|c|ccc|c}
    \hline
    \toprule
    \rowcolor{gray!20} Model& \textbf{Scene} & \textbf{Rec.} & \textbf{Gro.} & \textbf{Rel.} & \textbf{ReID} & \textbf{Sec.} & \textbf{psum} & \textbf{Loc.} & \textbf{A.C.} & \textbf{P.C.}  & \textbf{nsum} \\ \hline
    \multicolumn{12}{c}{\textbf{(a) Zero-shot}} \\
    \hline
    Pandagpt-7B & 28.49  & 45.10  & 23.22  & 47.75  & 23.69  & 22.10  & 190.35  & 39.54  & 25.19  & 78.29  & 143.02  \\
    LLaVA1.5-7B & 35.89  & 23.11  & 11.17  & 23.68  & 2.61  & 18.63  & 115.09  & 83.12  & 42.39  & 117.22  & 242.73  \\
    Qwen2-VL-7B & 49.80  & 59.42  & 25.74  & \underline{56.19}  & 4.62  & 30.08  & 225.85  & 50.41  & 18.05  & 89.12  & 157.58  \\
    InternVL2-8B & 46.02  & 77.69  & 30.05  & 53.02  & 7.33  & 38.19  & 252.30  & 43.29  & 21.75  & 60.39  & 125.43  \\
    InternVL2-26B & 62.24  & 72.88  & \textbf{39.48} & 55.42  & 8.16  & 43.72  & 281.90  & 38.25  & 21.68  & 64.65  & 124.58  \\ \hline
    IRGPT (anti-CL) & 61.29  & 80.25  & 27.89  & 52.66  & 19.44  & \underline{45.31}  & 286.84  & 35.22  & 20.29  & 53.93  & 109.44  \\
    IRGPT (Random) & \underline{63.93}  & \underline{83.38}  & 33.91  & 53.67  & \underline{28.18}  & 44.11  & \underline{307.18}  & \underline{34.99}  & \underline{15.33}  & \textbf{47.17} & \underline{97.49}   \\
    IRGPT (CL) & \textbf{65.12} & \textbf{86.28} & \underline{36.68}  & \textbf{58.33} & \textbf{33.55} & \textbf{48.69} & \textbf{328.65}  & \textbf{33.32} & \textbf{13.25} & \underline{47.30}  & \textbf{93.87} \\
       \hline
    \multicolumn{12}{c}{\textbf{(b) Fine-tune}} \\
    \hline
    Pandagpt-7B &  82.93 & 99.12 & 49.04 & 97.22 & 47.51 & 99.32 & 475.14 & 12.29 & 1.31  & 6.13  & 19.73  \\
    LLaVA1.5-7B & 80.04 & 96.32 & 44.48 & 95.11 & 42.89 & 97.56 & 456.40 & 20.69 & 1.71  & 11.66 & 34.06  \\
    Qwen2-VL-7B & 83.21 & 98.47 & 47.93 & 95.99 & 44.83 & 98.84 & 469.27 & 16.44 & 1.48  & 7.39  & 25.31  \\
    InternVL2-8B & 82.14  & 97.78  & 47.59  & 93.25  & 44.97  & 98.76  & 464.49  & 18.70   & 1.52  & 6.64  & 26.85  \\
    InternVL2-26B &  \underline{84.03}  & 98.94  & 48.77  & 95.89  & 47.09  & 99.06  & 473.78  & 12.50  & 1.38  & 5.68  & 19.57  \\
    \hline
    IRGPT (anti-CL) &  79.89  & 97.65  & 42.01  & 95.23  & 23.09  & 95.37  & 433.24  & 19.02  & 1.61  & 5.87  & 26.50  \\
    IRGPT (Random) &  83.89  & \underline{99.07}  & \underline{50.32}  & \underline{97.13}  & \underline{49.68}  & \underline{99.37}  & \underline{479.46}  & \underline{9.74}  & \underline{1.12}  & \underline{2.09}   & \underline{12.94}  \\
    IRGPT (CL)  & \textbf{85.12} & \textbf{99.79} & \textbf{51.58} & \textbf{98.69} & \textbf{50.79} & \textbf{99.82} & \textbf{485.79} & \textbf{3.32} & \textbf{0.25} & \textbf{0.82} & \textbf{4.39} \\
    \bottomrule \hline
    \end{tabular}}
  \caption{Comparison with other models. Among the 9 task comparisons, there are 6 positive metrics and 3 negative metrics, represented by positive sum (psum) and negative sum (nsum) respectively. The tasks are denoted by abbreviations with their corresponding full names in \cref{Benchmark}. We compared zero-shot and fine-tuned results. The difference is whether models were fine-tuned on our instruction dataset. The best results are highlighted in \textbf{bold}, and the sub-optimal results are \underline{underlined}.}
  \label{tab:main result}%
\end{table*}%

\subsubsection{Curriculum Schedule} 
The distance scores and alignment scores derived from the two lessons establish distinct sample rankings from their respective perspectives. We constructed a comprehensive ranking of all samples through systematic integration of these dual sequence rankings. However, the static nature of sample difficulty in this configuration might compromise generalization performance. Empirical insights \cite{zheng2023coverage,yang2024mind} suggest that introducing stochasticity to difficulty ordering can enhance model generalization bounds. Following the classical baby-step \cite{spitkovsky2010baby}, we partition the training data into $M$ tiers based on difficulty levels. During pre-training, tiers are sequentially ordered by difficulty while implementing random sampling within each tier.
This method achieves dual optimization objectives: 1) It preserves structured learning progression while introducing controlled variability in difficulty exposure, enhancing the model's adaptability to diverse challenge levels. 2) The stochastic tiered sampling provides inherent error tolerance for potential scoring discrepancies, creating more resilient learning dynamics compared to deterministic sequencing approaches.

Furthermore, we incorporate the variation rate derived from \cref{loss variation} as an adaptive weighting factor to simultaneously mitigate misaligned sample impacts and intensify focus on challenging instances. The final formulation integrates these components through:
\begin{equation}
\small
w_i= \begin{cases}1-\sigma\left(\frac{\alpha_i}{\operatorname{Median}\left(\left\{\alpha_j \mid \alpha_j>0\right\}\right)}\right) & \alpha_i>0, \\ 1+\sigma\left(\frac{-\alpha_i}{\operatorname{Median}\left(\left\{-\alpha_k \mid \alpha_k \leq 0\right\}\right)}\right) & \alpha_i \leq 0,\end{cases}\\
\end{equation}
\begin{equation}
\small
    \mathcal{L} = \frac{1}{N} \sum_{i=1}^N w_i \cdot \left[ -\sum_{c=1}^C y_{i,c} \log p_{i,c}\right],
\end{equation}
where Median$(\cdot)$ denotes the median operator and $\sigma(\cdot)$ represents the sigmoid function.

\section{Experiment}

\subsection{Experiment Setting}
\textbf{Implementation Details.}
We utilized InternVL2-8B as the base model and adopted the LoRA strategy for LLM fine-tuning. During the incremental pre-training phase, the learning rate was set to 1e-3 for 1 epoch, while the supervised instruction fine-tuning phase employed a learning rate of 2e-5 for 5 epochs, with a consistent batch size of 32 across both stages. We train our model with an AdamW \cite{loshchilovdecoupled} optimizer and a cosine learning rate scheduler.

\textbf{Evaluation Metrics.}
Given the diverse output requirements across the 9 infrared-text tasks, we implemented different evaluation metrics for each task's specific characteristics. For \textit{Location}, \textit{Aerial Counting}, and \textit{Pedestrian Counting} tasks, we calculated prediction errors using MAE. The \textit{Grounding} task involving bounding boxes was evaluated by mAP@0.5, while other tasks were measured using accuracy (Acc). For clarity, we aggregate the positive (psum) and negative (nsum) score metrics separately, where higher psum and lower nsum indicate better performance.

\textbf{Compared Methods.}
We selected five MLLMs for comparison: PandaGPT-7B \cite{su2023pandagpt} (a pioneer in adapting LLMs to infrared images), LLaVA1.5-7B \cite{LLaVA}, Qwen2-VL-7B \cite{wang2024qwen2}, InternVL2-8B \cite{chen2024internvl} (baseline), and InternVL2-26B (larger-scale variant). IRGPT comprises three versions with distinct training strategies: anti-CL (hard to easy), Random (full randomization), and CL (strategy defined in \cref{Curriculum Schedule}).

\subsection{Main Results}

Observed from \cref{tab:main result}, our IRGPT model pre-trained with bi-cross-modal curriculum transfer learning achieves state-of-the-art zero-shot performance on 7 tasks, outperforming the baseline InternVL2-8B by 76.35 psum, with a notable 26.22 improvement on ReID tasks. The 31.56 reduction in nsum suggests enhanced capability in object localization and counting.
After fine-tuning, our method surpasses all models including the larger InternVL2-26B, achieving the highest psum of 485.79 (12.01 over InternVL2-26B) with a nsum of merely 4.39 (15.18 lower than InternVL2-26B). 

Regarding training strategies, our bi-cross-modal curriculum transfer learning demonstrates clear superiority over random training, achieving a notable improvement of 14.47 in psum under zero-shot settings. However, we observe that the nsum metric exhibits a marginal decrease of merely 2.12, failing to demonstrate comparable improvement to psum. Notably, the Random variant achieves superior performance in personnel counting tasks.
We attribute this discrepancy to the pretraining phase's primary emphasis on image-text alignment, which facilitates object category and spatial relation identification, rather than precise image comprehension and task-specific responses (e.g., localization and object counting).
The anti-CL approach with inverse difficulty ordering proves particularly challenging, leading to severe performance degradation. For instance, in ReID tasks, the fine-tuned model frequently collapses to the same outputs, completely failing to capture pedestrian identity characteristics. More experiments and details are provided in the Appendix E$\&$I.

\subsection{Ablation Study}

\begin{table}[t]
  \centering
  \resizebox{0.8\linewidth}{!}{
    \begin{tabular}{ccccc|cr}
    \toprule
    \textbf{L1} & \textbf{anti-L1} & \textbf{L2} & \textbf{anti-L2} &\textbf{$\alpha$} & \textbf{psum} & \textbf{nsum} \\
    \hline
      &       &       &       &       &   464.49 & 26.85 \\
      & \checkmark &       & \checkmark & \checkmark & 433.24 & 26.50 \\
      &       &       & \checkmark & \checkmark &  454.73  & 24.36 \\
      &       & \checkmark &       & \checkmark & 470.23 & 17.29 \\
      & \checkmark &       &       & \checkmark & 456.30 & 24.59 \\
    \checkmark &       &       &       & \checkmark & 466.81 & 19.33 \\
    \checkmark &       & \checkmark &       &      &    481.36   & 7.08\\
    \hline
    \checkmark &       & \checkmark &       & \checkmark & 485.79 & 4.39 \\
    \bottomrule
    \end{tabular}}
  \caption{The ablation study on the pre-training effects of two cross-modal lessons. All comparisons here are based on results after instruction fine-tuning.}
  \label{tab:ablation1}%
\end{table}%

\begin{table}[t]
  \centering
  \resizebox{0.9\linewidth}{!}{
    \begin{tabular}{l|cc|cc}
    \toprule
    \multicolumn{1}{c|}{\multirow{2}[1]{*}{\textbf{Schedule}}} & \multicolumn{2}{c|}{\textbf{zero-shot}} & \multicolumn{2}{c}{\textbf{fine-tune}} \\ \cmidrule{2-5}    
          & \textbf{psum}  & \textbf{nsum}  & \textbf{psum}  & \textbf{nsum} \\ \midrule
    Difficulty-Ascending & 309.63  & 101.90  & 474.29  & 19.24  \\
    Difficulty-Descending &   286.84    &   109.44    &    433.24   & 26.50 \\
    Bidirectional Ordering & 308.29  & 105.08  & 469.84  & 19.53  \\
    Random & 307.18  & 97.49  & 479.46  & 12.94  \\
    Descending-Stratified Random &   302.85    &   103.62   &    470.22   & 18.92 \\
    Ascending-Stratified Random & 328.65  & 93.87  & 485.79  & 4.39  \\
\bottomrule   \end{tabular}
    }
  \caption{The ablation study of sampling schedule for batch selection. Difficulty-Descending/Ascending indicate difficulty-ordered selection. Bidirectional Ordering balances batch difficulty via alternating directions. `-Stratified' implies difficulty-based grouping. Ascending-Stratified Random is our method in \cref{Curriculum Schedule}.}
  \label{tab:schedule}%
\end{table}%
Ablation studies cover curriculum components and sampling strategies. The curriculum ablation results in \cref{tab:ablation1} demonstrate that removing curriculum and using anti-curriculum learning both degrade infrared image comprehension. Notably, Lesson2 (L2) shows greater sensitivity to psum than Lesson1 (L1), as L2 emphasizes image-text similarity whereas L1 prioritizes image quality. Image quality difficulty affects perceptual accuracy, while alignment complexity impacts semantic correctness.

As shown in \cref{tab:schedule}, we compare different curriculum schedules. Our findings reveal that all reverse difficulty-based strategies adversely affect performance, while approaches incorporating randomness consistently enhance model effectiveness. Notably, even the descending-stratified random strategy achieves significantly superior results compared to the difficulty-descending approach.

\subsection{Visualization}

\begin{figure}
    \centering
    \includegraphics[width=0.85\linewidth]{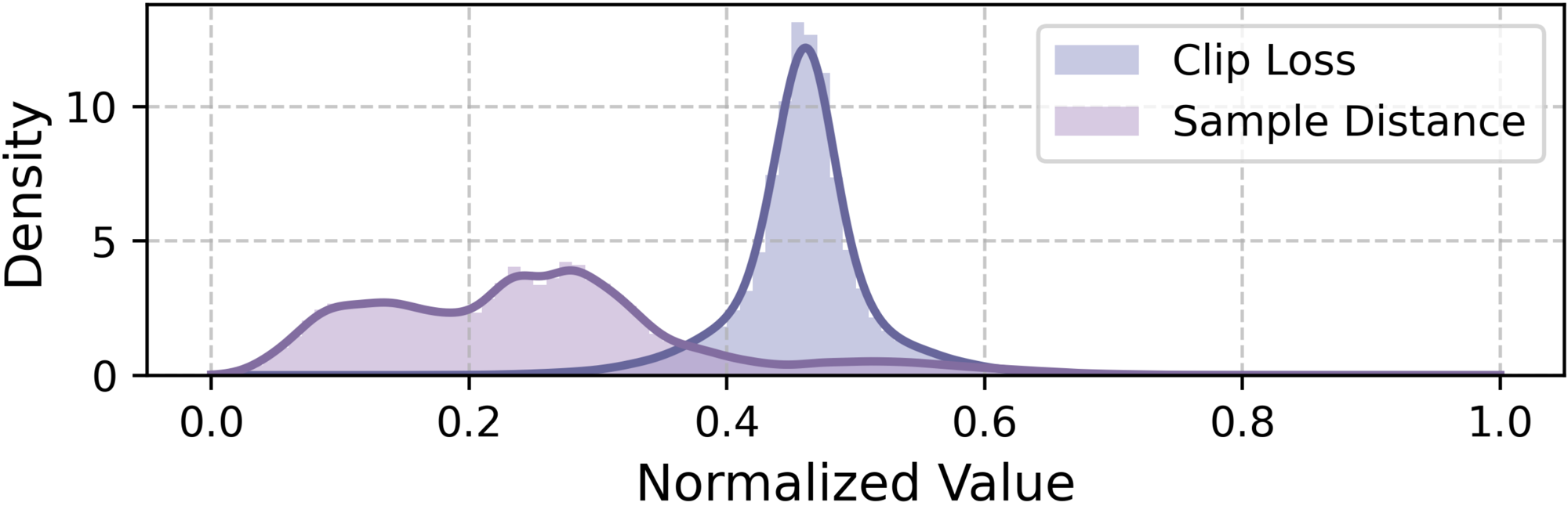}
    \caption{The distribution of difficulty score. The CLIP loss exhibits a normal distribution, while the sample distances show a bimodal distribution.}
    \label{fig:distribution}
\end{figure}

\textbf{Sample Difficulty Distribution.} 
We analyze normalized evaluation metric distributions from two experimental cases. As shown in \cref{fig:distribution}, the sample distance distribution exhibits a bimodal pattern, arising from infrared cameras' predominant use of two spectral bands (thermal imaging and near-infrared) coupled with significant clarity variations across datasets. CLIP's normally distributed losses stem from the Central Limit Theorem compliance, and infrared-text alignment noise follows Gaussian distributions. These statistical properties do not affect difficulty ranking validity, as our metric relies on relative score rankings instead of absolute values.


\begin{figure}
    \centering
    \includegraphics[width=0.94\linewidth]{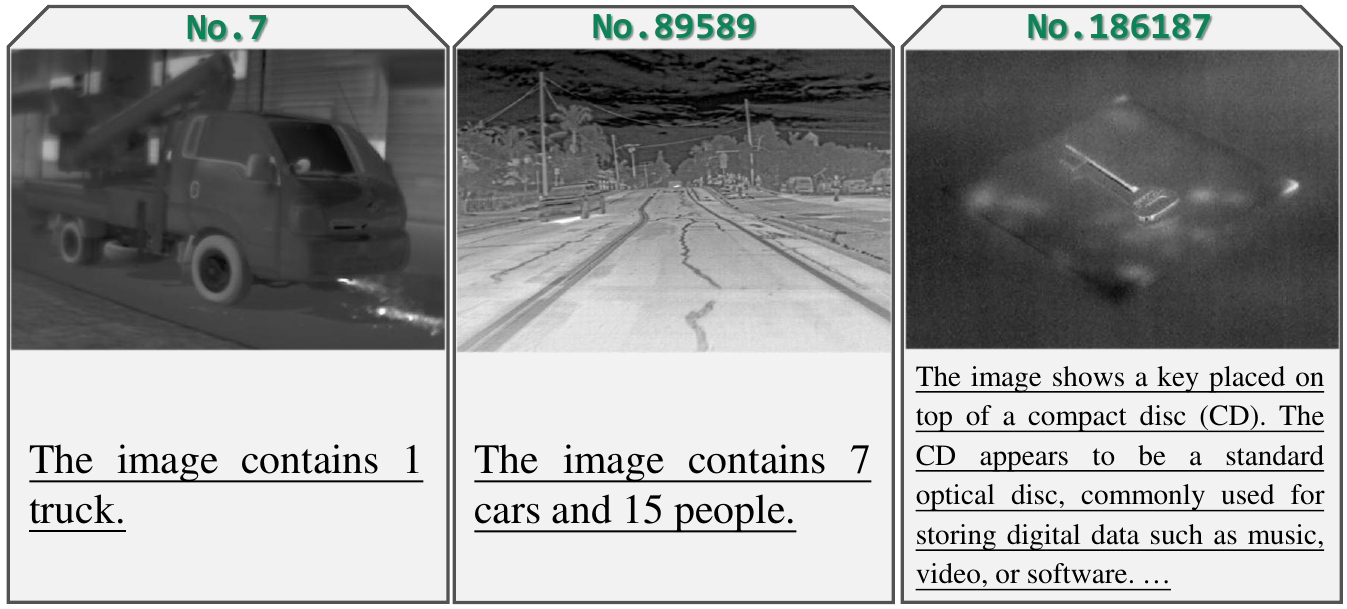 }
    \caption{Sample pairs with ranking results. Demonstrating variations in semantic complexity and image quality.}
    \label{fig:order}
\end{figure}

\textbf{Example of Difficulty Ranking.} \cref{fig:order} presents three illustrative samples that demonstrate discernible trends in image quality and text-image alignment. The learning difficulty of the samples exhibits a progressive escalation corresponding to ascending rankings. This empirical evidence substantiates the efficacy of the difficulty metric employed in our bi-cross-modal curriculum transfer learning framework, where sample complexity shows positive correlation with hierarchical ranking positions.

\section{Conclusion}

We introduce IRGPT, the first vision-language model specifically designed for real-world infrared image understanding. Through the development of the large-scale IR-TD dataset and a bi-cross-modal curriculum transfer learning framework, our approach successfully addresses the modality gap between visible and infrared domains while maintaining authentic semantic representation. The proposed model achieves state-of-the-art performance across multiple benchmarks, demonstrating superior efficiency compared to larger-scale counterparts. 

\section*{Acknowledge}
This work was funded by the STI 2030—Major Projects under grant 2022ZD0209600, the National Natural Science Foundation of China under grant 62201058, 62475016 and 62306311, and the Science and Technology on Electromechanical Dynamic Control Laboratory Funding under grant 6142601012402.
We also thank the anonymous reviewers for their constructive comments and Prof. Shuo Yang from Harbin Institute of Technology, Shenzhen, for valuable suggestions during the rebuttal period.